\documentclass[runningheads]{llncs}
\usepackage[T1]{fontenc}
\usepackage{graphicx,verbatim}
\usepackage{booktabs}
\usepackage{multirow}
\usepackage{graphicx}
\usepackage{subcaption}
\usepackage{amsmath}
\usepackage{enumitem}
\usepackage{adjustbox}
\usepackage{caption}
\usepackage{hyperref}
\usepackage{cleveref}
\usepackage{amsfonts}
\usepackage[table,xcdraw]{xcolor}
\usepackage{colortbl}
\usepackage{tikz}   
\usepackage[normalem]{ulem}
\useunder{\uline}{\ul}{}
\crefname{figure}{Fig.}{Figs.}
\crefname{table}{Tab.}{Tabs.}
\usepackage{color}

\definecolor{highcolor}{HTML}{137333} 
\definecolor{lowcolor}{HTML}{5353ec}
\definecolor{midcolor}{HTML}{FFFFFF} 
\newcommand{\mrbhldiff}[2]{%
  \begingroup
  \pgfmathsetmacro{\diffval}{#1 - #2}%
  \pgfmathsetmacro{\scaled}{abs(\diffval)^0.7 * 285}%
  \pgfmathtruncatemacro{\mrbint}{min(55, max(5, \scaled))}%
  \ifdim \diffval pt > 0pt
    \edef\mrbcol{highcolor!\mrbint!midcolor}%
    \expandafter\cellcolor\expandafter{\mrbcol}#1%
  \else
    \ifdim \diffval pt < 0pt
      \edef\mrbcol{lowcolor!\mrbint!midcolor}%
      \expandafter\cellcolor\expandafter{\mrbcol}#1%
    \else
      #1%
    \fi
  \fi
  \endgroup
}

\newcommand{\iconStar}{\raisebox{-0.2em}{\includegraphics[height=0.9em]{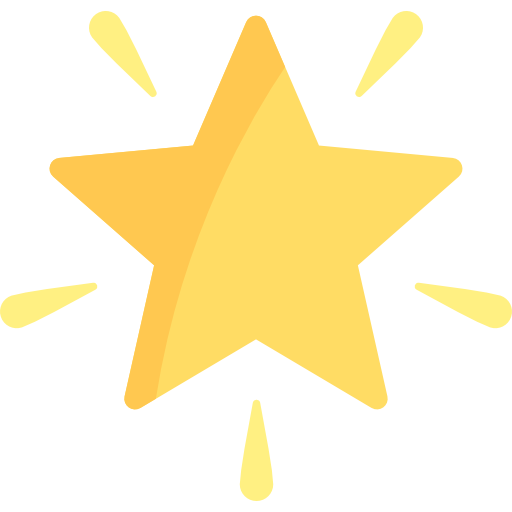}}}

\newif\ifcomments
\commentstrue  

\begin{document}
\title{Dementia-Agents: A Multi-Modal Multi-Agent System for Dementia Staging and Phenotyping}
\titlerunning{Dementia-Agents}
%

\author{
Yaling Shen*\textsuperscript{1}, 
Maja Christensen*\textsuperscript{1,2},
Yiwen Jiang\textsuperscript{1},
Jenna Dennison\textsuperscript{3},\\
David Darby\textsuperscript{1,2},
Amy Brodtmann\textsuperscript{1,2},
Zongyuan Ge\textsuperscript{1}
}  
\authorrunning{Shen et al.}
\institute{
\textsuperscript{1}Monash University, 
\textsuperscript{2}Eastern Health,
\textsuperscript{3}Lived Experience Advisor\\
    \email{\{yaling.shen, zongyuan.ge\}@monash.edu}}
  
\maketitle              
\begin{abstract}
Dementia diagnosis requires integrating multi-modal clinical assessments from diverse informants and clinicians under incomplete and heterogeneous data conditions.
Yet most AI-driven approaches remain Alzheimer’s disease (AD)-centric, framing the problem as binary AD detection or three-stage AD progression modeling within well-curated research settings. 
This pathology-driven paradigm overlooks the broader, syndrome-level nature of dementia, which spans multiple stages, phenotypes, and etiologies.
In this paper, we propose \textbf{Dementia-Agents}, a clinically aligned multi-agent framework for real-world dementia staging and phenotyping. 
The framework follows a three-step workflow: 
(1) a data agent translates structured clinical records into semantically faithful textual representations that preserve missing-data signals and routes them to domain-aligned experts;
(2) five fine-tuned expert agents generate domain-level predictions; and
(3) a coordinator agent performs probabilistic aggregation to produce final staging and phenotyping decisions.
We develop and evaluate Dementia-Agents on a real-world clinical cohort of 1,066 patients from two cognitive neurology services.
Compared with monolithic multi-modal large language models (MLLMs) and prior medical multi-agent systems, our approach achieves consistent improvements in diagnostic performance for real-world syndrome-level dementia staging and phenotyping, while preserving domain-level interpretability.


\end{abstract}
%
%
%
\begin{figure}[t]
\includegraphics[width=\textwidth]{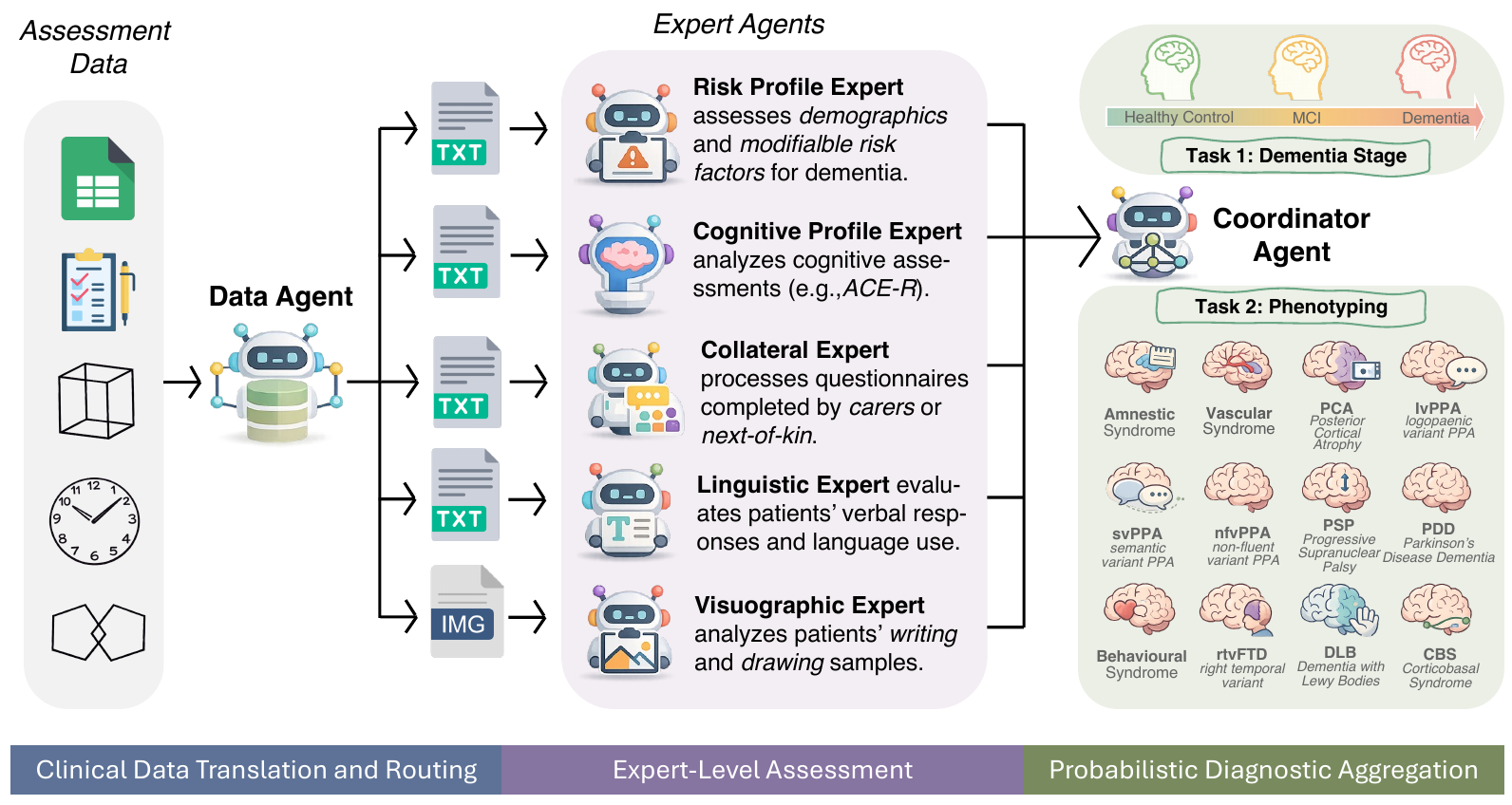}
\caption{The workflow of Dementia-Agents with three main steps.} 
\label{fig:teaser}
\end{figure}

\section{Introduction}
Dementia diagnosis represents a complex multi-modal process, requiring clinicians to interpret diverse assessment results from both patients and caregivers, including demographics, cognitive testing records (e.g., Mini-Mental State Examination (MMSE)~\cite{mmse}), and questionnaire-based reports~\cite{xue2024nm}.
Despite growing interest in AI-driven dementia research, much prior work has focused primarily on Alzheimer’s disease (AD) detection, typically framing the dementia diagnosis as a binary classification~\cite{care-ad,dementia-r1} or three-stage AD progression modeling~\cite{HuWen_AnatomyGuided_MICCAI2025,diamond} based on cognitive screening scores and neuroimaging data~\cite{adni}.
However, dementia is a heterogeneous clinical syndrome of progressive cognitive and functional decline, whereas AD is a specific neuropathological disease and the most common cause of dementia~\cite{elahi2017clinicopathological,arvanitakis2019diagnosis}. 
As such, dementia diagnosis extends beyond AD to involve multiple phenotypes and etiologies, necessitating syndrome-level staging and phenotyping rather than pathology-driven detection alone.

While recent advances in multi-modal large language models (MLLMs) have improved automated diagnosis systems~\cite{vlm4alz}, their adaptation to dementia remains largely AD-centric and grounded in well-curated experimental cohorts, limiting applicability to routine clinical settings.
Many systems rely on longitudinal datasets~\cite {dementia-r1,care-ad} or comprehensive multi-source cohorts~\cite{xue2024nm}, assuming complete data availability, fixed modality combinations, and biomarker-supported labels.
Such assumptions diverge from real-world dementia assessment, where data are heterogeneous, partially observed, informant-dependent, and collected incrementally across clinical visits, posing challenges for reliable modeling and deployment in routine practice.

The emergence of multi-agent frameworks enables task decomposition to simulate multi-disciplinary collaboration in the medical diagnostic pipeline~\cite{medagents,medagentpro}.
Dementia assessment data, however, presents a different form of complexity.
Rather than spanning multiple medical disciplines, it involves multi-party inputs from patient- and caregiver-reported responses, alongside assessments conducted by multiple clinicians.
This assessment structure is not explicitly modeled by existing medical multi-agent systems.
Although an agent-based system has been proposed for AD~\cite{adagent}, its scope remains limited to pathology-driven AD detection and does not extend to syndrome-level dementia diagnosis.
Consequently, dementia-specific multi-agent modeling aligned with real-world clinical workflows remains largely underexplored.

To address these dementia-specific challenges, we propose \textbf{Dementia-Agents}, a clinically aligned multi-agent framework for syndrome-level dementia staging and phenotyping.
The framework decomposes dementia assessment into clinically defined domains, enabling integration of multi-source, multi-modal evidence under incomplete data conditions.
As shown in \cref{fig:teaser}, it follows a three-step pipeline:
(1) \textbf{Clinical data translation and routing}, where a data agent converts structured clinical records into textual representations while preserving missing-data semantics and routes them to domain-specific experts;
(2) \textbf{Expert-level assessment}, in which five fine-tuned expert agents generate their respective domain-level diagnostic predictions; and
(3) \textbf{Probabilistic diagnostic aggregation}, where a coordinator agent integrates these predictions into final staging and phenotyping decisions.
The framework is developed and validated on a real-world cohort of 1,066 individuals from two cognitive neurology services collected between 2012 and 2024.

In summary, our contributions are threefold. 
First, we present Dementia-Agents, a clinically aligned multi-agent framework that models syndrome-level dementia diagnosis as a distributed process consistent with real-world workflow. 
Second, we propose a structured expert decomposition that enables domain-specific evidence routing, coordinated prediction under missing data, and domain-level interpretability. 
Third, we demonstrate consistent performance improvements over monolithic MLLMs and multi-agent designs on a real-world clinical cohort, supported by systematic ablation and domain-level weighting analyses.


\section{Dementia-Agents}
As illustrated in~\cref{fig:teaser}, our Demetia-Agents comprises three main steps: clinical data translation and routing, expert-level assessment, and probabilistic diagnostic aggregation.
We detail each of them below.


\subsection{Clinical Data Translation and Routing}
Our dataset comprises routinely collected clinical assessments, including demographics, modifiable risk factors, standardized cognitive tests, and caregiver-reported questionnaires, originally stored in tabular form. 
A \textbf{Data Agent} translates these records into semantically organized textual representations. 
For each domain, item-level outcomes are grouped according to their recorded status (e.g., present, absent, correct, incorrect, or unknown) while preserving task descriptions, responses, numerical scores, and missing-value indicators.
The resulting texts are then routed to five clinically defined expert agents. 

\noindent\textbf{- Risk Profile Expert.} 
This agent receives summaries of demographic variables (e.g., ethnicity, sex) and nine modifiable dementia risk factors spanning vascular and non-vascular categories (e.g., hypertension, diabetes, depression). 

\noindent\textbf{- Cognitive Profile Expert.}
This expert models the serialized ACE-R assessment across \textit{orientation and attention} (e.g., serial subtraction and backward spelling), \textit{memory} (e.g., anterograde and retrograde memory), \textit{language} (e.g., repetition and writing), \textit{verbal fluency} (e.g., phonemic and semantic), and \textit{visuospatial abilities} (e.g., figure copying and clock drawing). 
Its input captures the full textualized task-response-score triad for each item. 
For fluency and visuographic tasks, only speech-transcribed responses and image-based content are routed to modality-specific agents, while task prompts and scores are retained.
The assessment yields a total score out of 100 with five section-level subscores.

\noindent\textbf{- Collateral Expert.}
This agent processes four caregiver-reported questionnaires, including the Cambridge Behavioral Inventory-Revised (CBI-R~\cite{cbi-r}), the BEHAVE-AD~\cite{behave-ad} behavioral scale, a functional assessment inventory, and a structured memory questionnaire.
Each item is expressed as a textual statement describing symptom presence, absence, or unknown status, with original numeric ratings mapped to degree-based severity descriptors (e.g., mild, moderate, severe) and duration indicators when available.
The resulting narrative captures behavioral change, psychiatric symptoms, and functional decline from the caregiver perspective while preserving missing-data semantics.

\noindent\textbf{- Linguistic Expert.}
This agent analyzes speech-derived ACE-R fluency tasks, including category (e.g., naming animals) and letter fluency (e.g., words beginning with ``\textit{P}'') under timed conditions. 
Its inputs include task descriptions, word sequences, fluency scores, and clinically defined speech annotations.

\noindent\textbf{- Visuographic Expert.}
This agent processes image-based ACE-R tasks, including figure copying, clock drawing, and sentence writing. 
Its inputs include the task description, reference stimulus (when applicable), the digitized patient drawing or handwriting, and the associated score statement.

\begin{figure}[t]
\includegraphics[width=\textwidth]{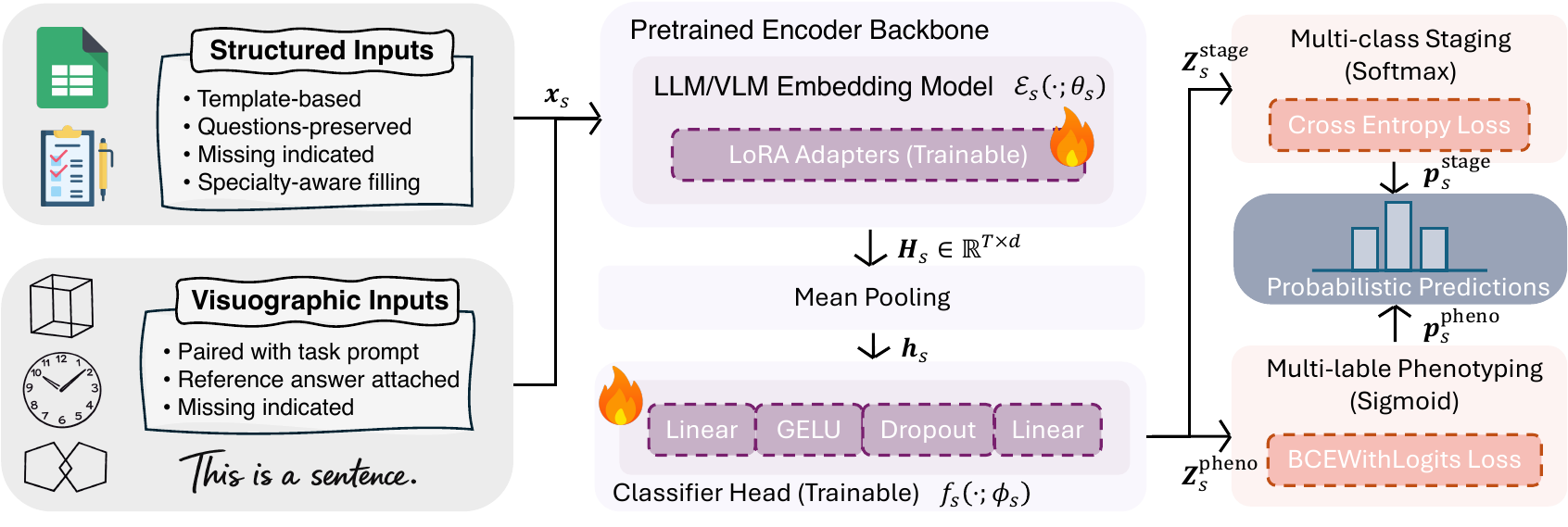}
\caption{The model architecture of each expert agent.} 
\label{fig:model}
\end{figure}
\subsection{Expert-Level Assessment}
\textbf{Task Formulation.}
Given domain-specific input $\mathbf{x}_s$ routed by the data agent, each expert $s$ applies a unified encoder-classifier pipeline to perform two prediction tasks: staging and phenotyping.
A pretrained encoder $\mathcal{E}_s(\cdot;\theta_s)$ produces token representations $\mathbf{H}_s \in \mathbb{R}^{T \times d}$, which are mean-pooled to obtain a fixed-length embedding $\mathbf{h}_s$ and projected to logits $\mathbf{z}_s$ via a two-layer MLP $f_s(\cdot;\phi_s)$:
\begin{equation}
\mathbf{h}_s = \text{MeanPool}(\mathbf{H}_s), \quad
\mathbf{z}_s = \mathbf{W}_{2,s}\,\text{GELU}(\mathbf{W}_{1,s} \mathbf{h}_s),
\end{equation}
where $\phi_s=\{\mathbf{W}_{1,s},\mathbf{W}_{2,s}\}$.
The logits are subsequently processed with task-specific output functions and losses for staging and phenotyping, respectively.

\textit{Staging.}
Dementia staging is formulated as a multi-class classification with ground-truth labels $y \in \{1,\dots, C\}$, where $C=3$. 
Applying softmax to logits $\mathbf{z}_s^{\text{stage}}$ yields class probabilities $\mathbf{p}_s^{\text{stage}}$. 
We minimize class-balanced cross-entropy:
\begin{equation}
\mathcal{L}_s^{\text{stage}} 
= - \omega_y \log p^{\text{stage}}_{s,y}, 
\end{equation}
where $\omega_y$ denotes the class weight and 
$p_{s,y}^{\text{stage}}$ the predicted probability of the ground-truth class $y$.

\textit{Phenotyping.}
Phenotype prediction is formulated as multi-label classification over $K=12$ categories (\cref{fig:teaser}), where the ground-truth label is a binary vector 
$\mathbf{y} \in \{0,1\}^K$.
Applying sigmoid to logits $\mathbf{z}_s^{\text{pheno}}$ yields label-wise probabilities $\mathbf{p}_s^{\text{pheno}}$, where $y_k$ and $p_{s,k}^{\text{pheno}}$ denote the $k$-th components. 
We minimize weighted binary cross-entropy with $\omega_k^{+}$ the positive-class weight for phenotype $k$:
\begin{equation}
    \mathcal{L}_s^{\text{pheno}} 
= - \sum_{k=1}^{K} 
\left[ \omega_k^{+} y_k \log p_{s,k}^{\text{pheno}}
+ (1-y_k) \log(1-p_{s,k}^{\text{pheno}}) \right],
\end{equation}
Each expert is optimized independently by minimizing its task-specific loss with respect to encoder parameters $\theta_s$ and MLP parameters $\phi_s$.

\subsection{Probabilistic Diagnostic Aggregation}
Given expert probability outputs $\mathbf{p}_s$ from the five expert agents, the \textbf{coordinator agent} computes a weighted aggregation with coefficients $\{\alpha_s\}$:
\begin{equation}
\mathbf{p} = \sum_{s} \alpha_s \mathbf{p}_s,
\quad 
\alpha_s \ge 0, \ \sum_s \alpha_s = 1.
\end{equation}
where $\{\alpha_s\}$ are optimized on the validation set via diverse strategies, including Bayesian optimization~\cite{bayesian}, derivative-free search (e.g., Nelder-Mead~\cite{nelder-mead}), evolutionary algorithms (e.g., Differential Evolution~\cite{differential}), and gradient-based methods, to obtain an optimal weight configuration, with the resulting coefficients tending to reflect relative diagnostic contributions across experts.

\begin{figure}[t]
\includegraphics[width=\textwidth]{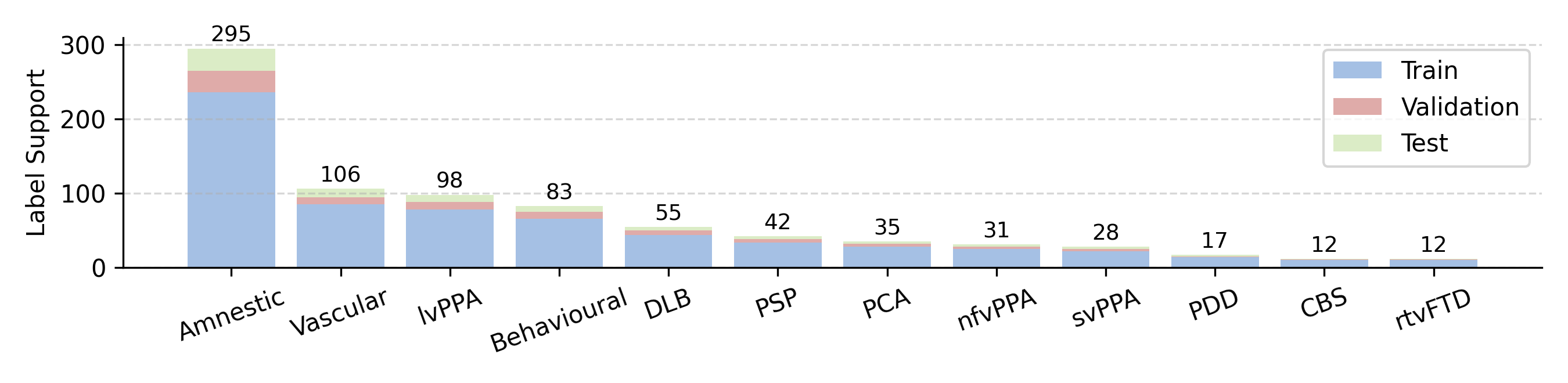}
\caption{Phenotype label support across training, validation, and test splits. The long-tailed distribution reflects real-world prevalence imbalance.} 
\label{fig:pheno-label}
\end{figure}

\section{Experimental Details}
\subsection{Dataset}
\noindent\textbf{Overview.}
Data were collected from two cognitive neurology services between 2012 and 2024, yielding a private cohort of 1,066 patients (mean symptom-onset age 63.3 years; 57\% male). 
The dataset is divided into training, validation, and test sets using stratified sampling to preserve label distributions. 
The held-out validation set is used by the coordinator agent for weight optimization, while experts are trained on the training set with internal validation splits.

\noindent\textbf{Staging.} 
All 1,066 samples are used for three-stage classification into healthy control, mild cognitive impairment (MCI), and dementia, with approximately balanced splits (Train: 275/283/302; Val: 35/29/39; Test: 28/34/41).

\noindent\textbf{Phenotyping.}
Excluding healthy controls, 722 samples are used for multi-label phenotyping. 
As illustrated in~\cref{fig:pheno-label}, stratified splitting preserves label ratios and ensures at least one positive instance per phenotype in each subset. 

\subsection{Experiment Setup}
\noindent\textbf{Baselines.}
Monolithic MLLM baselines include general-purpose Qwen3-VL-8B~\cite{qwen3technicalreport} and InternVL3.5-8B~\cite{internvl3_5}, as well as medical-specific Hulu-Med-7B~\cite{hulu-med} and LLaVA-Med-7B~\cite{llavam-med}, evaluated consistently across settings.
For multi-agent comparison, we implement MedAgents~\cite{medagents}, MedAgentPro~\cite{medagentpro}, and a zero-shot variant of Dementia-Agents, all built on Qwen3-VL-8B.

\noindent\textbf{Metrics.}
For staging, we report accuracy, macro-AUC, and macro-F1 to evaluate balanced performance across severity levels.
For phenotyping, we report macro-F1 and macro-AUC to account for label imbalance, micro-F1 for overall performance, and Hit@3 (i.e., whether the ground-truth phenotype appears within the top-3 predictions) to reflect clinical differential diagnosis relevance.

\noindent\textbf{Implementation Details.}
Multi-modal experts use Qwen3-VL-Embedding-8B~\cite{qwen3vlembedding}, while text-only ones use Qwen3-Embedding-8B~\cite{qwen3embedding} as encoders. 
Each expert is trained independently with a task-specific MLP (hidden size 2048), adapting the encoder using LoRA~\cite{lora} for 20 epochs with a batch size of 8.
Distinct learning rates are applied to the encoder and MLP head, tuned per expert. 
All experiments are conducted on a single 80GB NVIDIA A100 GPU. 
\begin{table}[t]
\centering
\caption{Performance comparison on multi-class \textbf{staging} and multi-label \textbf{phenotyping} under different settings. \textbf{Bold} indicates the best result and \underline{underlined} the second-best in each column.}
\label{tab:main}
\scalebox{0.8}{
\begin{tabular}{@{}crccccccc@{}}
\toprule
\multicolumn{2}{l}{} & \multicolumn{3}{c}{\textbf{Staging}} & \multicolumn{4}{c}{\textbf{Phenotyping}} \\ \cmidrule(lr){3-5} \cmidrule(lr){6-9}
\multicolumn{2}{l}{\multirow{-2}{*}{}} & AUC$_\text{Macro}$ & Acc. & F1$_\text{Macro}$ & AUC$_\text{Macro}$ & F1$_\text{Macro}$ & F1$_\text{Micro}$ & Hit@3 \\ \midrule
 & \multicolumn{8}{c}{\cellcolor[HTML]{E8E8E8}\textbf{Zero-Shot}} \\
 & Qwen3-VL-8B~\cite{qwen3technicalreport} & 0.528 & 0.417 & 0.261 & 0.501 & 0.109 & 0.145 & 0.319 \\
 & InternVL3.5-8B~\cite{internvl3_5} & 0.689 & 0.456 & 0.316 & 0.510 & 0.068 & 0.222 & 0.435 \\
 & Hulu-Med-7B~\cite{hulu-med} & 0.584 & 0.417 & 0.323 & 0.468 & 0.134 & 0.149 & 0.174 \\
 & LLaVA-Med-7B~\cite{llavam-med} & 0.369 & 0.350 & 0.176 & 0.576 & 0.168 & 0.186 & 0.435 \\ 
 & \multicolumn{8}{c}{\cellcolor[HTML]{E8E8E8}\textbf{Few-Shot}} \\ 
 & Qwen3-VL-8B~\cite{qwen3technicalreport} & 0.722 & 0.602 & 0.605 & \multicolumn{4}{l}{} \\
 & InternVL3.5-8B~\cite{internvl3_5} & 0.477 & 0.408 & 0.215 & \multicolumn{4}{l}{} \\
 & Hulu-Med-7B~\cite{hulu-med} & 0.578 & 0.398 & 0.190 & \multicolumn{4}{l}{} \\
 & LLaVA-Med-7B~\cite{llavam-med} & 0.439 & 0.398 & 0.190 & \multicolumn{4}{l}{\multirow{-4}{*}{\begin{tabular}[c]{@{}c@{}}* Three examples, one per label,\\are provided. Hulu-Med and \\LLaVA-Med predict all samples\\as \textit{``Dementia''}.\end{tabular}}} \\
 & \multicolumn{8}{c}{\cellcolor[HTML]{E8E8E8}\textbf{Supervised Fine-Tuning}} \\ 
 & Qwen3-VL-8B~\cite{qwen3technicalreport} & 0.821 & 0.670 & 0.658 & 0.615 & 0.149 & 0.249 & {\ul 0.667} \\
 & InternVL3.5-8B~\cite{internvl3_5} & 0.787 & 0.641 & 0.594 & 0.631 & 0.156 & 0.260 & 0.638 \\
 & LLaVA-Med-7B~\cite{llavam-med} & {\ul 0.841} & {\ul 0.689} & {\ul 0.680} & {\ul 0.637} & 0.161 & 0.254 & 0.609 \\
\multirow{-15}{*}{\begin{tabular}[c]{@{}c@{}}Single \\ MLLM\end{tabular}} & Hulu-Med-7B~\cite{hulu-med} & 0.814 & 0.650 & 0.630 & 0.616 & 0.163 & 0.255 & {\ul 0.667} \\ \midrule
 & MedAgents~\cite{medagents} & 0.718 & 0.505 & 0.431 & 0.508 & {\ul 0.190} & 0.219 & 0.594 \\
 & MedAgentPro~\cite{medagentpro} & 0.500 & 0.330 & 0.165 & 0.525 & 0.121 & {\ul 0.308} & 0.551 \\
 & Ours (Zero-Shot) & 0.637 & 0.476 & 0.362 & 0.505 & 0.179 & 0.210 & 0.362 \\
\multirow{-4}{*}{\begin{tabular}[c]{@{}c@{}}Multi- \\ Agent\end{tabular}} & \iconStar\textbf{Ours} & \textbf{0.874} & \textbf{0.757} & \textbf{0.745} & \textbf{0.743} & \textbf{0.242} & \textbf{0.364} & \textbf{0.754} \\ \bottomrule
\end{tabular}
}
\caption{Leave-one-out ablation on \textbf{both} tasks. \textcolor{blue}{Color} intensity indicates the magnitude of performance change from the full model.}
\label{tab:leave-one-out}
\scalebox{0.8}{
\begin{tabular}{@{}lccccccc@{}}
\toprule
 & \multicolumn{3}{c}{\textbf{Staging}} & \multicolumn{4}{c}{\textbf{Phenotyping}} \\ \cmidrule(lr){2-4} \cmidrule(lr){5-8}
\multirow{-2}{*}{} & AUC$_\text{Macro}$ & Acc. & F1$_\text{Macro}$ & AUC$_\text{Macro}$ & F1$_\text{Macro}$ & F1$_\text{Micro}$ & Hit@3 \\ \midrule
\multicolumn{1}{c}{\textbf{Ours}} & \textbf{0.874} & \textbf{0.757} & \textbf{0.745} & \textbf{0.743} & \textbf{0.242} & \textbf{0.364} & \textbf{0.754} \\ \midrule
w/o Risk Profile & \mrbhldiff{0.869}{0.874} & \mrbhldiff{0.738}{0.757}  & \mrbhldiff{0.724}{0.745} & \mrbhldiff{0.738}{0.743} & \mrbhldiff{0.207}{0.242}  & \mrbhldiff{0.313}{0.364} & \mrbhldiff{0.725}{0.754} \\
w/o Cognitive & \mrbhldiff{0.829}{0.874} & \mrbhldiff{0.709}{0.757} & \mrbhldiff{0.694}{0.745} & \mrbhldiff{0.725}{0.743} & \mrbhldiff{0.207}{0.242} & \mrbhldiff{0.319}{0.364} & \mrbhldiff{0.696}{0.754} \\
w/o Collateral & \mrbhldiff{0.860}{0.874} & \mrbhldiff{0.748}{0.757} & \mrbhldiff{0.739}{0.745} & \mrbhldiff{0.680}{0.743} & \mrbhldiff{0.192}{0.242} & \mrbhldiff{0.239}{0.364} & \mrbhldiff{0.725}{0.754} \\
w/o Linguistic & \mrbhldiff{0.870}{0.874} & \mrbhldiff{0.728}{0.757} & \mrbhldiff{0.707}{0.745} & \mrbhldiff{0.728}{0.743} & \mrbhldiff{0.208}{0.242} & \mrbhldiff{0.305}{0.364} & \mrbhldiff{0.710}{0.754} \\
w/o Visuographic & \mrbhldiff{0.863}{0.874} & \mrbhldiff{0.728}{0.757} & \mrbhldiff{0.717}{0.745} & \mrbhldiff{0.711}{0.743} & \mrbhldiff{0.231}{0.242} & \mrbhldiff{0.299}{0.364} & \mrbhldiff{0.725}{0.754} \\
w/o Coordinator & \mrbhldiff{0.867}{0.874} & \mrbhldiff{0.738}{0.757} & \mrbhldiff{0.725}{0.745} & \mrbhldiff{0.725}{0.743} & \mrbhldiff{0.211}{0.242} & \mrbhldiff{0.309}{0.364} & \mrbhldiff{0.710}{0.754} \\
\bottomrule
\end{tabular}
}
\end{table}

\section{Results and Analyses}
\noindent\textbf{Main Results.}
\cref{tab:main} compares Dementia-Agents against (i) monolithic MLLMs under zero-shot, few-shot, and supervised fine-tuning settings, and (ii) alternative medical multi-agent designs, for both multi-class staging and multi-label phenotyping.
In the \textit{zero-} and \textit{few-shot settings}, monolithic MLLMs exhibit limited performance, particularly for phenotyping.
For staging, several models degenerate to predicting all cases as \textit{``Dementia''}, revealing a dementia-biased prior induced by lexical cues in assessment narratives.
\textit{supervised training} improves all baselines, yet Dementia-Agents consistently achieves the strongest results. 
Compared with the best supervised monolithic MLLM baseline, we improve staging F1 by 8.7\% and obtain 28.6\% relative gains in phenotyping micro-F1. 
Dementia-Agents also outperforms prior multi-agent designs, indicating that clinically-aligned expert decomposition provides a more appropriate inductive structure for modeling the multi-source dementia assessment.

\noindent\textbf{Ablation Study.}
\cref{tab:leave-one-out} reports leave-one-out ablations across expert agents. 
Removing any expert consistently degrades performance, confirming that each contributes non-redundant clinical signals. 
Staging performance declines most when the \textit{Cognitive} expert is removed, reflecting the central role of direct cognitive testing in severity discrimination.
In contrast, phenotyping is most sensitive to exclusion of the \textit{Collateral} expert, underscoring the importance of caregiver-reported behavioral patterns in syndrome differentiation. 
Replacing the \textit{Coordinator} with uniform weighting further reduces performance, demonstrating that expert contributions are not equally informative. 
Overall, phenotyping exhibits greater sensitivity to ablation, suggesting that fine-grained syndrome modeling depends more critically on clinically aligned expert decomposition.

\begin{figure*}[t]
\centering
\begin{subfigure}[t]{0.32\textwidth}
    \centering
    \includegraphics[width=\linewidth]{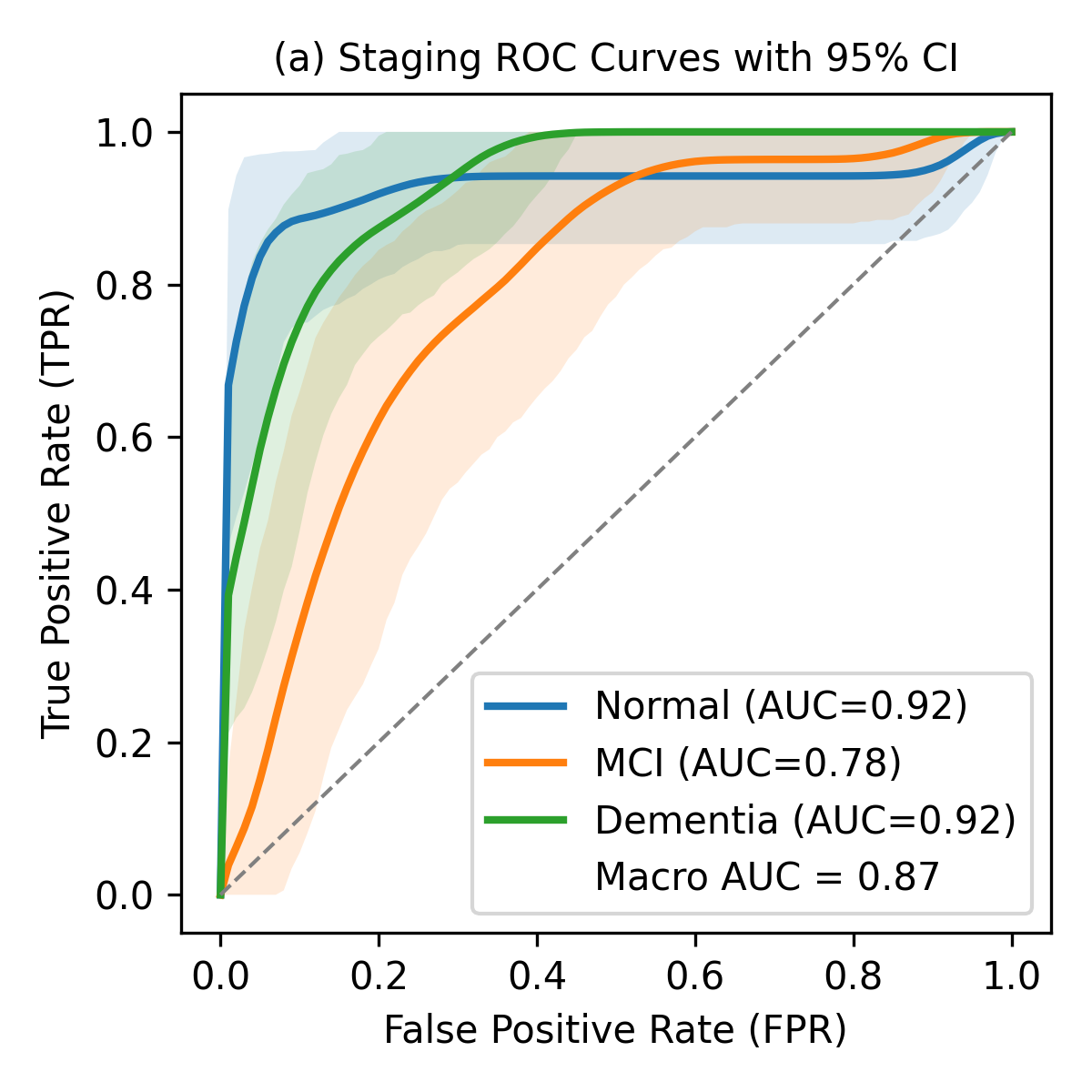}
    \label{fig:stage_roc}
\end{subfigure}
\hfill
\begin{subfigure}[t]{0.32\textwidth}
    \centering
    \includegraphics[width=\linewidth]{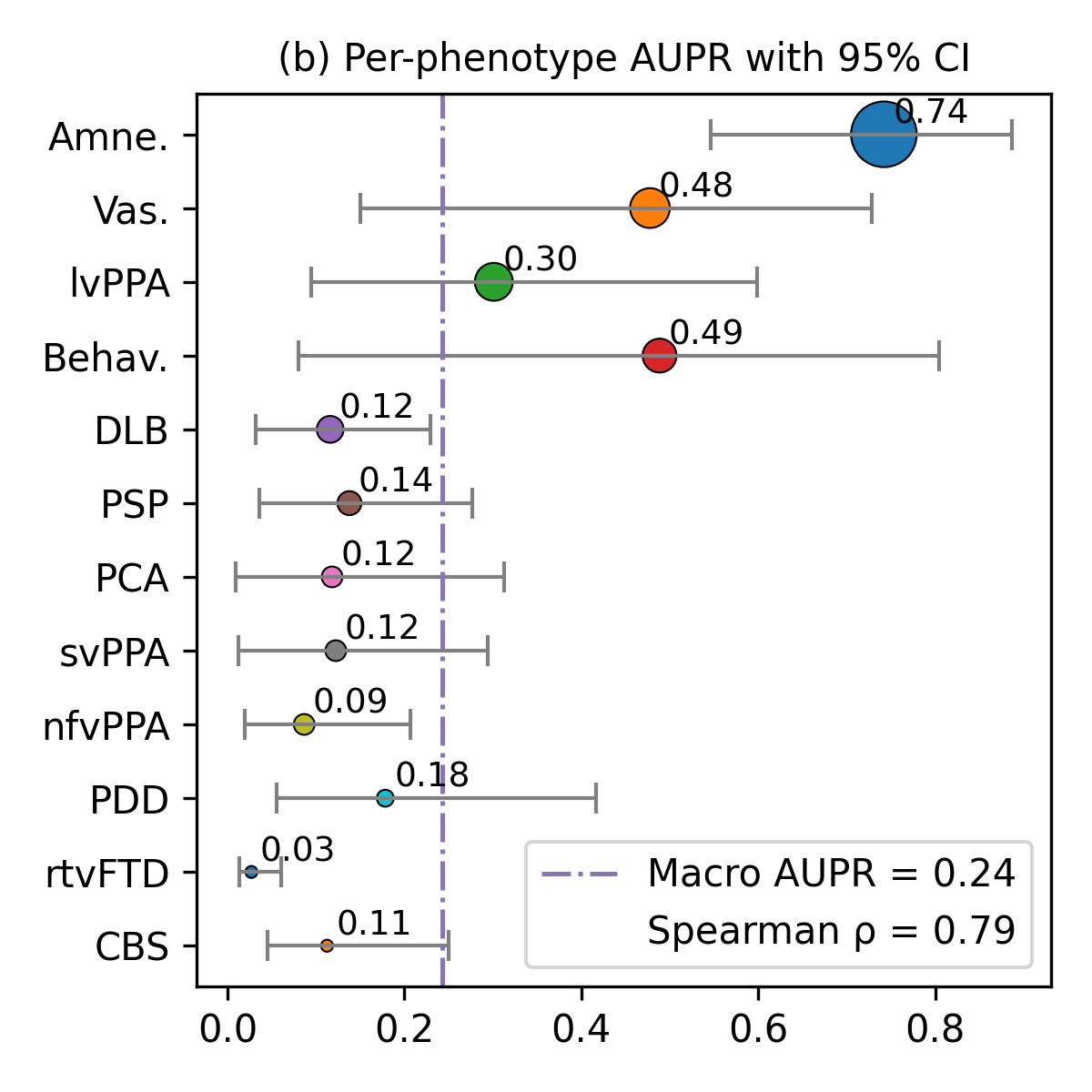}
    \label{fig:pheno_aupr}
\end{subfigure}
\hfill
\begin{subfigure}[t]{0.32\textwidth}
    \centering
    \includegraphics[width=\linewidth]{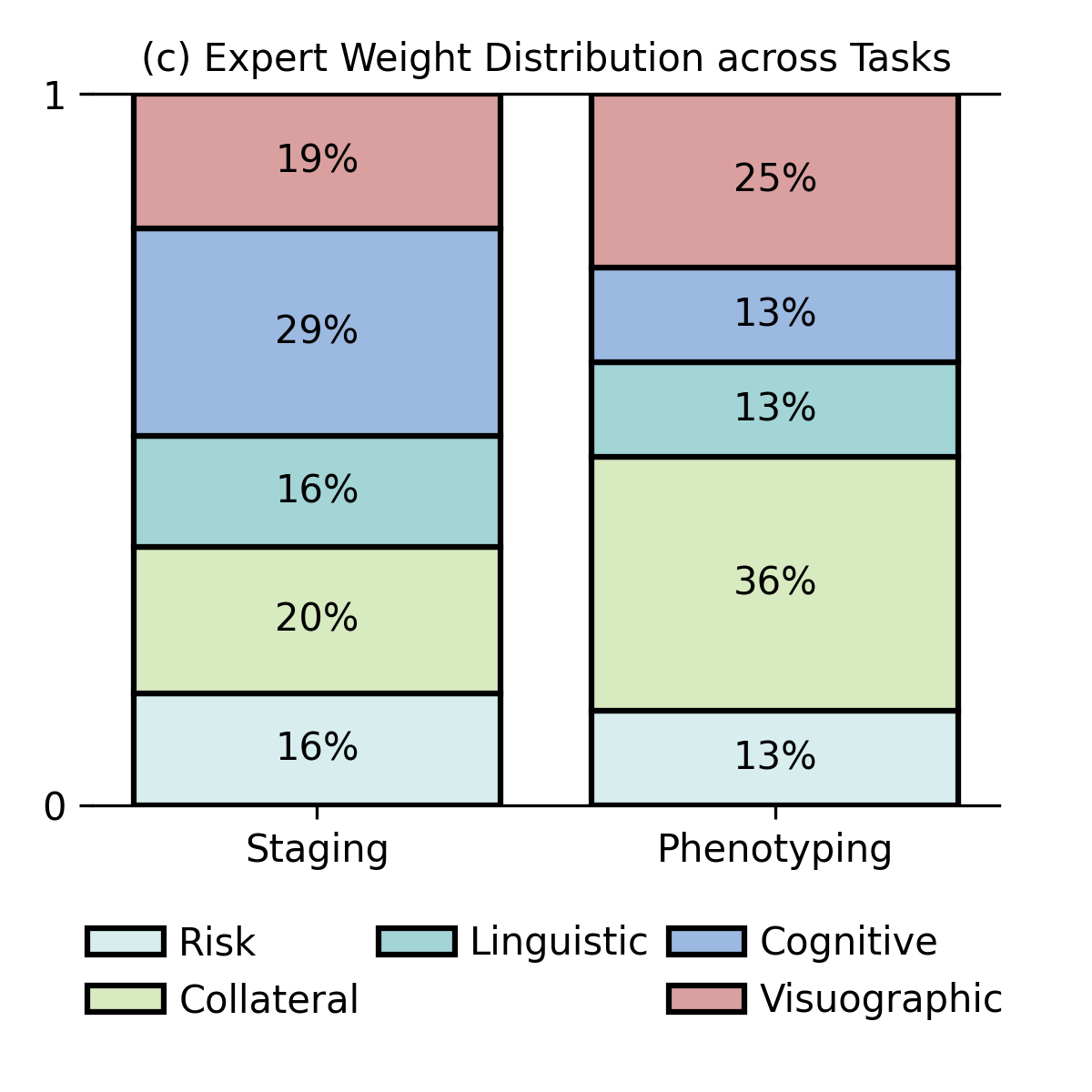}
    \label{fig:weights}
\end{subfigure}
\caption{Diagnostic performance and aggregation analysis.}
\label{fig:analysis}
\end{figure*}

\noindent\textbf{Performance Analysis.}
\cref{fig:analysis}(a) shows staging ROC curves with 95\% bootstrap confidence intervals (CI). 
While overall discrimination is strong, MCI exhibits comparatively lower separability, reflecting its intermediate clinical presentation. 
\cref{fig:analysis}(b) reports per-phenotype AUPR with 95\% CI. 
Performance correlates positively with label prevalence (Spearman $\rho=0.79$), aligning with the long-tailed distribution of dementia subtypes.
These results indicate that the proposed framework maintains stable performance across both transitional severity boundaries and imbalanced subtype distributions.

\noindent\textbf{Expert Weighting.}
\cref{fig:analysis}(c) visualizes the expert weights optimized by the \textit{Coordinator} agent. 
Staging assigns greater weight to \textit{Cognitive} evidence, whereas phenotyping emphasizes \textit{Collateral} information. 
These task-dependent weighting patterns indicate that expert contributions vary according to diagnostic objectives rather than following a fixed hierarchy. 
Independent clinician review indicated that the learned distributions are consistent with routine diagnostic processes, supporting the clinical plausibility of the aggregation strategy.           


\section{Conclusion}
We present Dementia-Agents, a multi-modal multi-agent framework for clinically aligned dementia staging and phenotyping on real-world assessments.
By routing heterogeneous inputs to expert agents and aggregating their probabilistic outputs, the framework improves diagnostic accuracy while maintaining an interpretable decision structure. 
Beyond pathology-driven AD detection, it supports syndrome-level dementia staging and phenotyping. 
Extensive experiments show consistent gains over monolithic MLLMs and medical multi-agent frameworks, with ablations highlighting the value of structured expert decomposition.


%
%
%
\bibliographystyle{splncs04}
\bibliography{mybibliography}
%




\end{document}